\pdfoutput=1

\documentclass[11pt]{article}

\usepackage[]{ACL2023}
\usepackage{graphicx}
\usepackage{times}
\usepackage{latexsym}

\usepackage[T1]{fontenc}

\usepackage[utf8]{inputenc}
\newcommand*{\affmark}[1][*]{\textsuperscript{#1}}

\usepackage{microtype}
\usepackage{subfigure}
\usepackage{inconsolata}

\usepackage[prependcaption,textsize=tiny]{todonotes}
\definecolor{ao}{rgb}{0.0, 0.5, 0.0}
\definecolor{dark}{rgb}{0.20784313725,0.10980392156,0.45882352941}
\definecolor{med}{rgb}{0.40392156862,0.30588235294,0.65490196078}
\definecolor{light}{rgb}{0.70588235294,0.65490196078,0.83921568627}
\definecolor{light_green}{rgb}{0.85098039215,0.91764705882,0.82745098039}
\definecolor{special_red}{rgb}{0.59607843137,0.59607843137,0.59607843137}
\definecolor{g1}{rgb}{0.36078431372,0.65098039215,0.2862745098}
\definecolor{g2}{rgb}{0.26666666666,0.50588235294,0.55294117647}

\title{SanskritShala: A Neural Sanskrit NLP Toolkit with Web-Based Interface for Pedagogical and Annotation Purposes}

\author{Jivnesh Sandhan\affmark[1], Anshul Agarwal\affmark[1], Laxmidhar Behera\affmark[1,3],\\ \textbf{Tushar Sandhan\affmark[1] and Pawan Goyal\affmark[2]}\\
\affmark[1]IIT Kanpur, \affmark[2]IIT Kharagpur, \affmark[3]IIT Mandi\\
\texttt{jivnesh@iitk.ac.in,pawang@cse.iitkgp.ac.in}}

\begin{document}
\maketitle
\begin{abstract}
We present  a  neural  Sanskrit  Natural Language Processing (NLP)  toolkit named SanskritShala\footnote{It means `a school of Sanskrit'.} to facilitate computational linguistic analyses for several tasks such as word segmentation, morphological tagging, dependency parsing, and compound type identification. Our systems currently report state-of-the-art performance on available benchmark datasets for all tasks. SanskritShala is deployed as a web-based application, which allows a user to get real-time analysis for the given input. It is built with easy-to-use interactive data annotation features that allow annotators to correct the system predictions when it makes mistakes. We publicly release the source codes of the 4 modules included in the toolkit, 7 word embedding models that have been trained on publicly available Sanskrit corpora and multiple annotated datasets such as word similarity, relatedness, categorization, analogy prediction to assess intrinsic properties of word embeddings.  So far as we know, this is the first neural-based Sanskrit NLP toolkit that has a web-based interface and a number of NLP modules. We are sure that the people who are willing to work with Sanskrit will find it useful for pedagogical and annotative purposes. SanskritShala is available at: \url{https://cnerg.iitkgp.ac.in/sanskritshala}. The demo video of our platform can be accessed at: \url{https://youtu.be/x0X31Y9k0mw4}.

\end{abstract}

\section{Introduction}
Sanskrit is a culture-bearing and knowledge-preserving language of ancient India. Digitization has come a long way, making it easy for people to access ancient Sanskrit manuscripts \cite{goyal-etal-2012-distributed,adiga-etal-2021-automatic}. 
However, we find that the utility of these digitized manuscripts is limited due to the  user's lack of language expertise and various linguistic phenomena exhibited by the language.
This motivates us to investigate how we can utilize natural language technologies to make Sanskrit texts more accessible.

The aim of this research is to create neural-based Sanskrit NLP systems that are accessible through a user-friendly web interface. The Sanskrit language presents a range of challenges for building deep learning solutions, such as the \textit{sandhi} phenomenon, a rich morphology, frequent compounding, flexible word order, and limited resources \cite{translist,Graph-Based,sandhan-etal-2021-little,sandhan-etal-2019-revisiting}. To overcome these challenges, 4 preliminary tasks were identified as essential for processing Sanskrit texts: word segmentation, morphological tagging, dependency parsing, and compound type identification. The word segmentation task is complicated by the \textit{sandhi} phenomenon, which transforms the word boundaries \cite{translist}. The lack of robust morphological analyzers makes it challenging to extract morphological information, which is crucial for dependency parsing. Similarly, dependency information is essential for several downstream tasks such as word order linearisation \cite{krishna-etal-2019-poetry} which helps to decode possible interpretation of the poetic composition. Additionally, the ubiquitous nature of compounding in Sanskrit is difficult due to the implicitly encoded semantic relationship between its constituents \cite{sandhan-etal-2022-novel}. 
These 4 tasks can be viewed as a preliminary requirement for developing robust NLP technology for Sanskrit.  
Thus, we develop novel neural-based linguistically informed architectures for all 4 tasks, reporting state-of-the-art performance on Sanskrit benchmark datasets \cite{sandhan-etal-2022-novel,translist,sandhan_systematic}. 

In this work, we introduce  a  neural  Sanskrit  NLP toolkit named SanskritShala\footnote{Roughly, it can be translated as `a school of Sanskrit'.} to assist computational linguistic analyses involving multiple tasks such as word segmentation, morphological tagging, dependency parsing, and compound type identification.
SanskritShala is also deployed as a web application that enables users to input text and gain real-time linguistic analysis from our pretrained systems. It is also equipped with user-friendly interactive data annotation capabilities that allow annotators to rectify the system when it makes errors. It provides the following benefits: (1) A user with no prior experience with deep learning can utilise it for educational purposes. (2) It can function as a semi-supervised annotation tool that requires human oversight for erroneous corrections.   
We publicly release the source code of the 4 modules included in the toolkit, 7 word embedding models that have been trained on publicly available Sanskrit corpora and multiple annotated datasets such as word similarity, relatedness, categorization, analogy prediction to measure the word embeddings' quality.
 To the best of our knowledge, this is the first neural-based Sanskrit NLP toolkit that contains a variety of NLP modules integrated with a web-based interface.

Summarily, our key contributions are as follows:
\begin{itemize}
    \item We introduce the first neural Sanskrit NLP toolkit to facilitate automatic linguistic analyses for 4  downstream tasks (\S \ref{neural_toolkit}).
    \item We release 7 pretrained Sanskrit embeddings and suit of 4 intrinsic evaluation datasets to measure the word embeddings' quality (\S \ref{embed_dataset}).
    \item We integrate SanskritShala with a user-friendly web-based interface which is helpful for pedagogical purposes and in developing annotated datasets (\S \ref{webinterface}).
    \item We publicly release codebase and datasets of all the modules of SanskritShala which currently mark the state-of-the-art results.\footnote{\url{https://github.com/Jivnesh/SanskritShala}}
\end{itemize}

\section{Related Work on Sanskrit NLP Tools}
Recently, the Sanskrit Computational Linguistics (SCL) field has seen significant growth in building web-based tools to help understand Sanskrit texts.
\newcite{goyal2016} introduced the Sanskrit Heritage Reader (SHR), a lexicon-driven shallow parser that aids in the selection of segmentation solutions.
Sams\={a}dhan{\=\i} is another web-based tool consisting of various rule-based modules \cite{kulkarni-sharma-2019-paninian,kulkarni-etal-2020-dependency,sriram-etal-2023-validation}.
Recently, \newcite{Sangrahaka,Chandojnanam} introduced a web-based annotation tool for knowledge-graph construction and a metrical analysis.

\begin{figure*}[!tbh]
    \centering
    \subfigure[]{\includegraphics[width=0.45\textwidth]{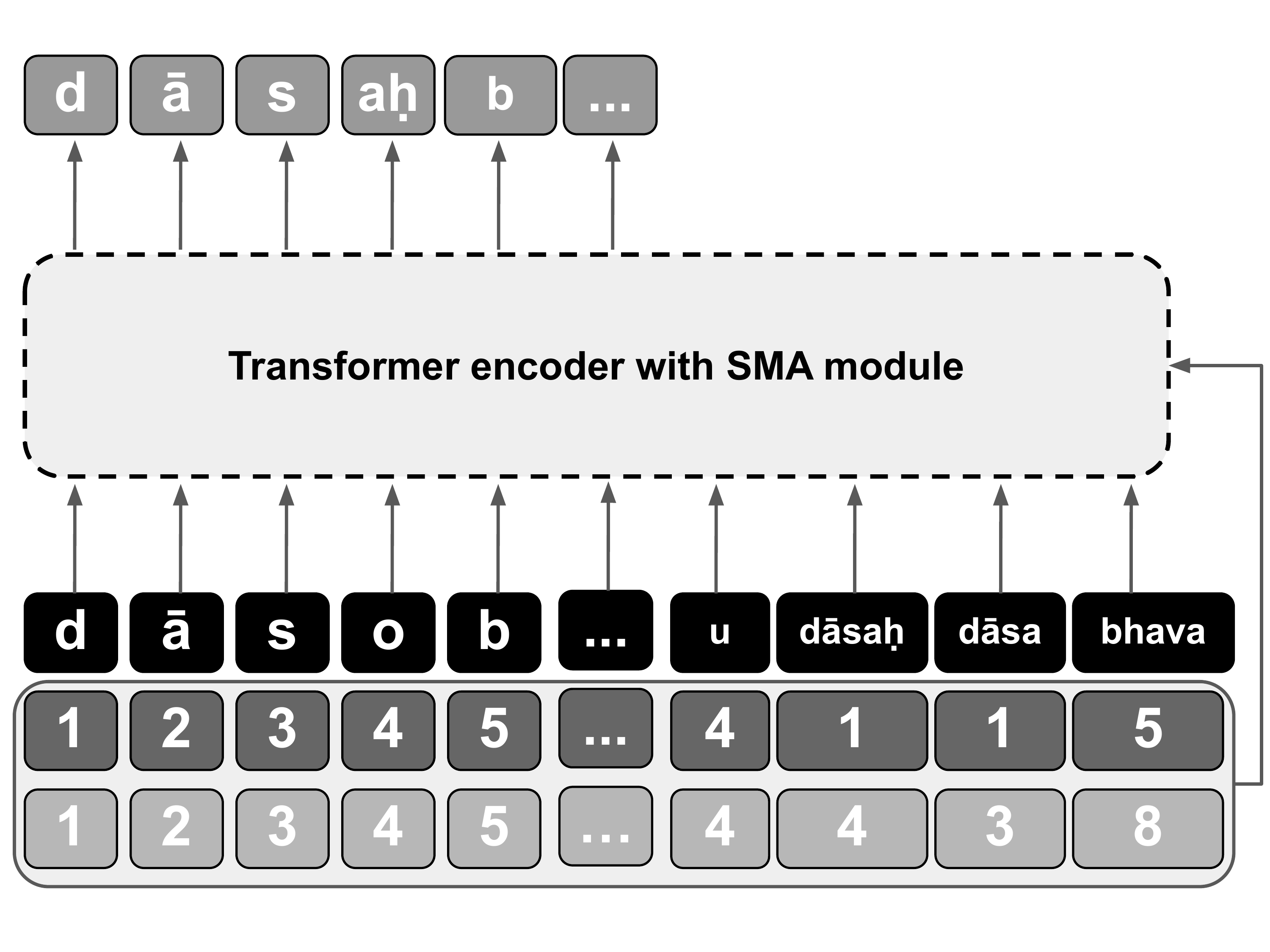}\label{fig:flat_architecture}}
    \subfigure[]
    {\includegraphics[width=0.45\textwidth]{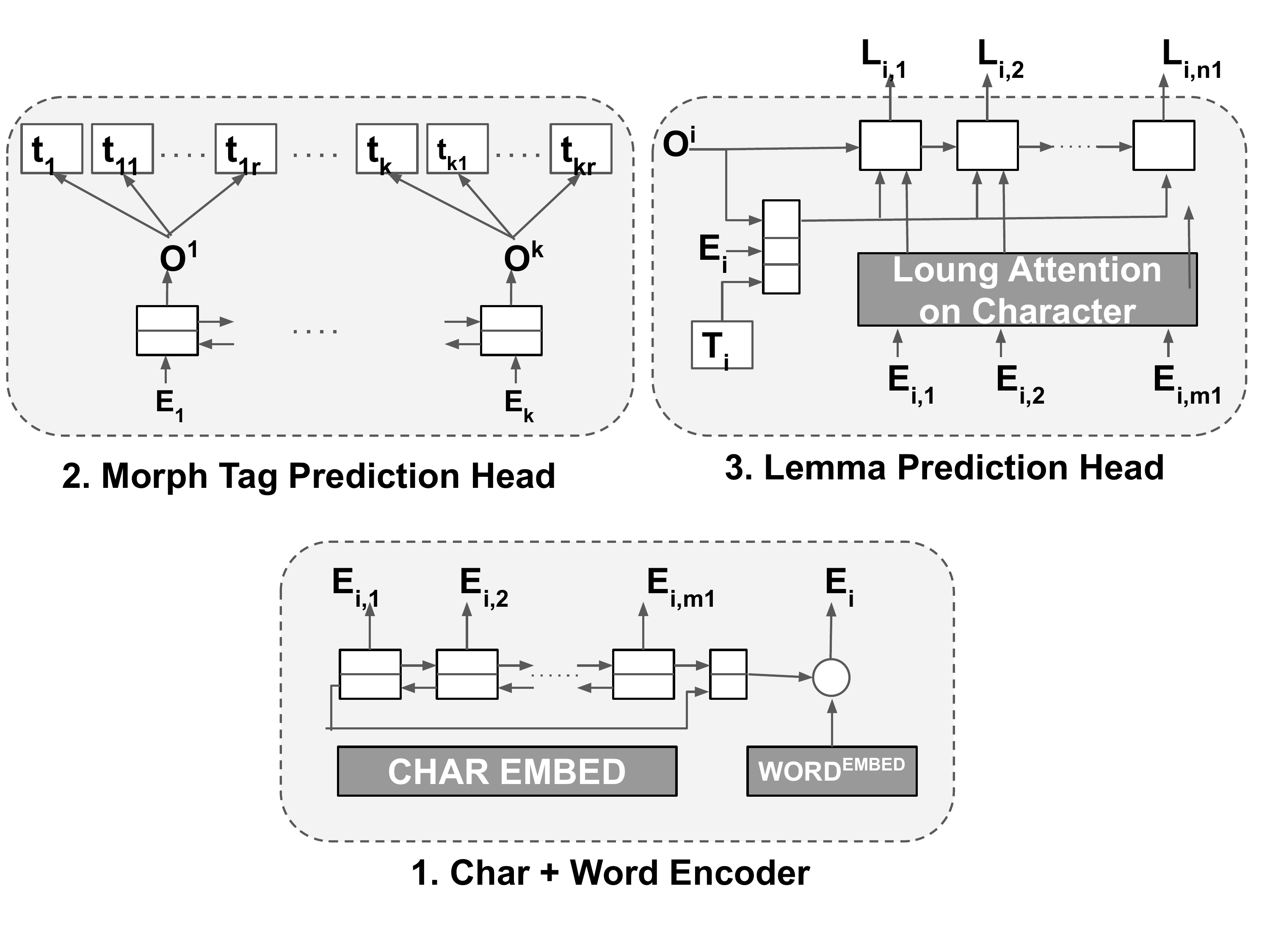}\label{fig:LemmaTag}}
    \subfigure[]
    {\includegraphics[width=0.45\textwidth]{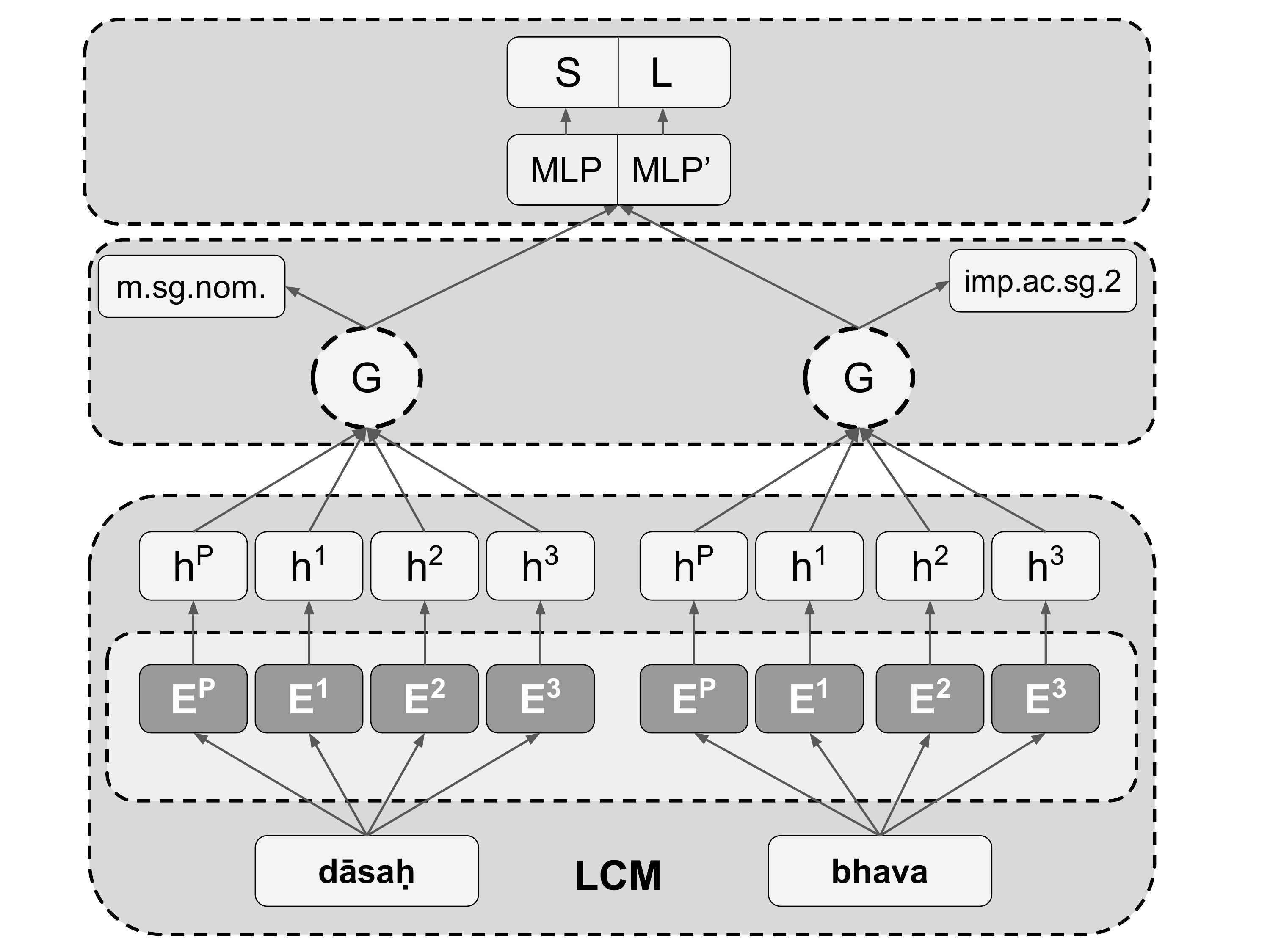}\label{fig:dep_architecture}}
    \subfigure[]
    {\includegraphics[width=0.45\textwidth]{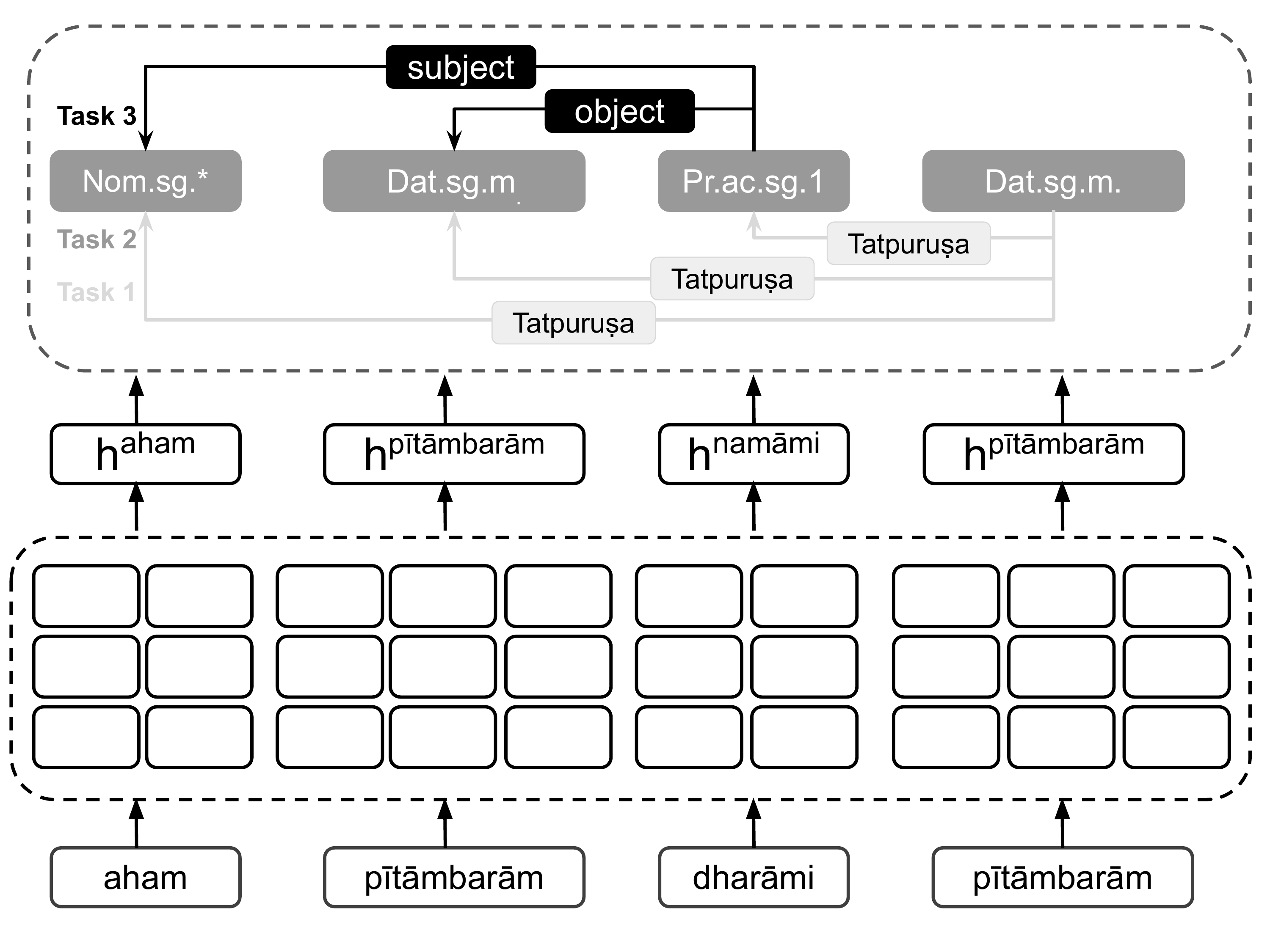}\label{fig:sacti_architecture}}
    \caption{(a) Toy illustration of the TransLIST system.``{\sl d\={a}sobhava}''. Translation: ``Become a servant.'' (b) LemmaTag architecture in which multi-task learning formulation is leveraged to predict morphological tags and lemmas by employing bidirectional RNNs with character-level and word-level representations. (c) Proposed ensembled architecture for dependency parsing integrated with the LCM pretraining. LCM is acronym for three auxiliary tasks: Lemma prediction, Case prediction and Morphological tag prediction.   (d) Toy example illustrating the context-sensitive multi-task learning system: ``{\sl aham p\={\i}ta-ambaram dhar\={a}mi}'' (Translation: ``I wear a yellow cloth'') where `{\sl p\={\i}ta-ambaram}' is a compound having \textit{Tatpuru\d{s}a} semantic class according to the context presented.}
    \label{figure:complete_TransLIST}
\end{figure*}

In short, tools for NLP can be divided into two groups: rule-based and annotation tools. Rule-based tools have limitations such as not providing a final solution, limited vocabulary coverage, and lacking user-friendly annotation features. Annotation tools, on the other hand, do not have the recommendations of rule-based systems, relying solely on annotators. To address these limitations, a web-based annotation framework called SHR++ \cite{krishna-etal-2020-shr} was proposed. It combines the strengths of both types of tools by offering all possible solutions from rule-based system SHR for tasks like word segmentation and morphological tagging, allowing annotators to choose the best solution rather than starting from scratch.

Our proposal, SanskritShala, goes a step further by integrating a neural-based NLP toolkit that combines state-of-the-art neural-based pre-trained models with rule-based suggestions through a web-based interface. Each module of SanskritShala is trained to predict the solutions from the exhaustive candidate solution space generated by rule-based systems. Hence, it makes predictions in real time using neural-based models that have already been trained. Thus, a complete solution is shown to the users / annotators, which was not possible in any of the previous attempts.

Further, annotators can easily correct the mispredictions of the system with the help of user-friendly web-based interface. This would significantly reduce the overall cognitive load of the annotators. To the best of our knowledge, SanskritShala is the first NLP toolkit available for a range of tasks with a user friendly annotation interface integrated with the neural-based modules.  

\section{About Sanskrit}
Sanskrit is an ancient language known for its cultural significance and knowledge preservation. However, it presents challenges for deep learning due to its morphological complexity, compounding, free word order, and lack of resources. Sanskrit's intricate grammar, with its combination of roots, prefixes, and suffixes, requires advanced algorithms to analyze and understand. Compounding adds another layer of complexity as multiple words combine to form new words with unique meanings \cite{krishna-etal-2016-compound,sandhan-etal-2022-novel}. The free word order in Sanskrit complicates tasks like parsing and understanding, requiring models to comprehend meaning regardless of word placement \cite{krishna-etal-2023-neural,krishna-etal-2019-poetry}. Moreover, Sanskrit is considered a low-resource language, lacking extensive datasets and pre-trained models \cite{sandhan-etal-2021-little}. Overcoming these challenges necessitates linguistic expertise, computational techniques, and sufficient language resources. Developing specialized models to handle Sanskrit's morphology, compounding, and word order is essential. Creating annotated datasets, lexicons, and corpora will also contribute to advancing research and applications in Sanskrit \cite{sandhan-etal-2022-prabhupadavani,sandhan-etal-2023-evaluating}. Despite the obstacles, utilizing deep learning to explore Sanskrit benefits the preservation of cultural heritage and facilitates a deeper understanding of India's literature and philosophy, while also pushing the boundaries of natural language processing.

\section{A Neural NLP Sanskrit Toolkit}
In this section, we describe SanskritShala, which is a neural Sanskrit NLP toolkit designed to aid computational linguistic analysis including various tasks, such as word segmentation, morphological tagging, dependency parsing, and compound type identification. It is also available as a web application that allows users to input text and obtain real-time linguistic analysis from our pretrained algorithms.
We elucidate SanskritShala by first elaborating on
its key modules.
\label{neural_toolkit}
\paragraph{Word Tokenizer:}
Earlier \textit{lexicon-driven} systems for Sanskrit word segmentation (SWS) rely on Sanskrit Heritage Reader \cite[SHR]{goyaldesign16}, a rule-based system, to obtain the exhaustive solution space for segmentation, followed by diverse approaches to find the most valid solution. However, these systems are rendered moot while stumbling out-of-vocabulary words. Later, \textit{data-driven} systems for SWS are built using the most recent techniques in deep learning, but can not utilize the available candidate solution space.  To overcome the drawbacks of both lines of modelling, we build a \textbf{Tran}sformer-based \textbf{L}inguistically-\textbf{I}nformed \textbf{S}anskrit \textbf{T}okenizer (TransLIST) \cite{translist} containing (1) a component that encodes the character-level and word-level potential candidate solutions, which tackles \textit{sandhi} scenario typical to SWS and is compatible with partially available candidate solution space, (2) a novel soft-masked attention for prioritizing selected set of candidates and (3) a novel path ranking module to correct the mispredictions. 
Figure \ref{fig:flat_architecture}  illustrates the TransLIST architecture, where the candidate solutions obtained from SHR are used as auxiliary information. 
In terms of the perfect match (PM) metric, TransLIST outperforms with 93.97 PM compared to the state-of-the-art \cite{hellwig-nehrdich-2018-sanskrit}  with 87.08 PM.
\paragraph{Morphological Tagger:}
Sanskrit is a morphologically-rich fusional Indian language with 40,000 possible labels for inflectional morphology \cite{krishna-etal-2020-graph,gupta-etal-2020-evaluating}, where homonymy and syncretism are predominant \cite{krishna-etal-2018-free}. We train a neural-based architecture \cite[LemmaTag]{kondratyuk-etal-2018-lemmatag} on Sanskrit dataset \cite{hackathon}. Figure \ref{fig:LemmaTag} illustrates the system architecture in which multi-task learning formulation is leveraged to predict morphological tags and lemmas by employing bidirectional RNNs with character-level and word-level representations.
Our system trained on the Sanskrit dataset stands first with 69.3 F1-score compared to the second position with 69.1 F1-score on the Hackathon dataset \cite{hackathon} leaderboard.\footnote{Hackathon leaderboard: \url{https://competitions.codalab.org/competitions/35744\#results}} 
\paragraph{Dependency Parser:}
Due to labelled data bottleneck, we focus on low-resource techniques for Sanskrit dependency parsing. Numerous strategies are tailored to improve task-specific performance in low-resource scenarios.
 Although these strategies are well-known to the NLP community, it is not obvious to choose the best-performing ensemble of these methods for a low-resource language of interest, and not much effort has been given to gauging the usefulness of these methods. We investigate 5 low-resource strategies in our ensembled Sanskrit parser \cite{sandhan_systematic}: data augmentation, multi-task learning, sequential transfer learning, pretraining, cross/mono-lingual and self-training.
Figure \ref{fig:dep_architecture} shows our ensembled system, which supersedes with 88.67 Unlabelled Attached Score (UAS) compared to the state-of-the-art \cite{krishna-etal-2020-keep} with 87.46 UAS for Sanskrit and shows on par performance in terms of Labelled Attached Score.
\paragraph{Sanskrit Compound Type Identifier (SaCTI)}
 is a multi-class classification task that identifies semantic relationships between the components of a compound. Prior methods only used the lexical information from the constituents and did not take into account the most crucial syntactic and contextual information for SaCTI. However, the SaCTI task is difficult mainly due to the implicitly encrypted context-dependent semantic relationship between the compound's constituents. 
Thus, we introduce a novel multi-task learning approach \cite{sandhan-etal-2022-novel} (Figure \ref{fig:sacti_architecture}) which includes contextual information and enhances the complementary syntactic information employing morphological parsing and dependency parsing as two auxiliary tasks. SaCTI outperforms with  $81.7$ F1-score compared to the state-of-the-art by \newcite{krishna-etal-2016-compound} with  $74.0$ F1-score. 
\begin{figure*}[!tbh]
    \centering
    \includegraphics[width=0.9\textwidth]{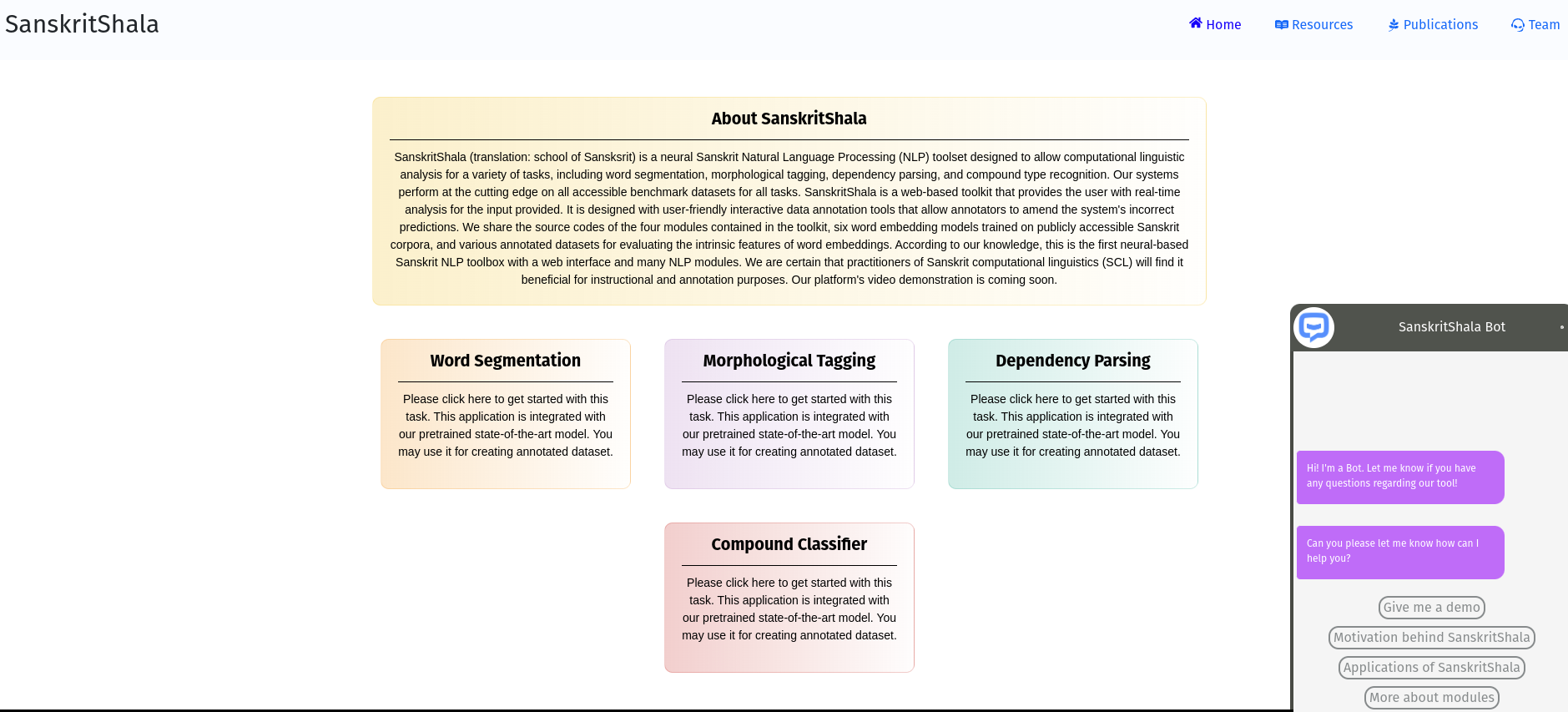}
    \caption{The web interface of the SanskritShala. 
    At the bottom right, a rule-based chatbot is added to navigate users on the platform to give users a user-friendly experience.}
    \label{figure:sanskritshala}
\end{figure*}
\begin{figure*}[!tbh]
    \centering
       \subfigure[]{\includegraphics[width=0.45\textwidth]{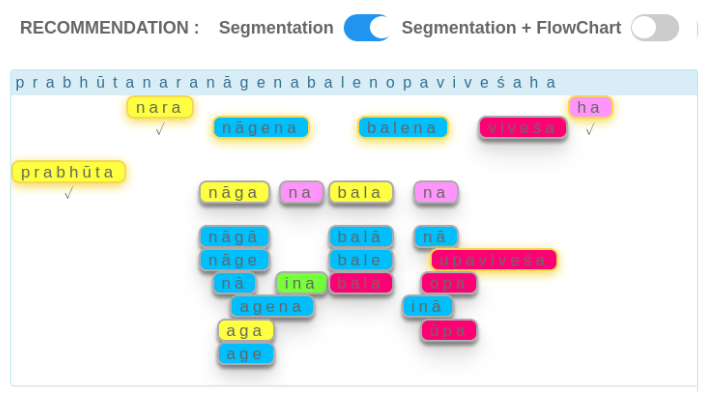}\label{fig:shr++}} 
    \subfigure[]{\includegraphics[width=0.45\textwidth]{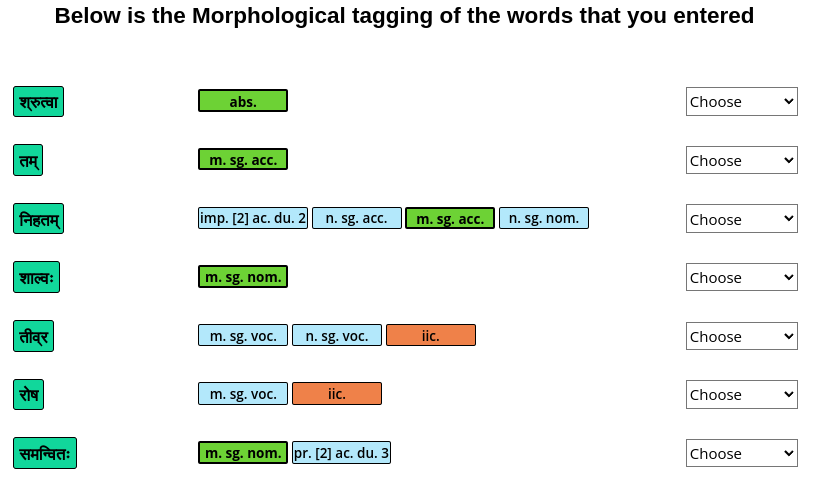}\label{fig:morph_tagger}}
    \caption{(a) The candidate solution space generated by SHR for the word segmentation task and the predicted solution by our pretrained model is recommended for the sequence ‘\textit{prabh\={u}tanaran\={a}gena balenopavive\'sa ha}’ using a yellow highlight. (b) Morphological Tagger: For each word, we show possible morphological analyses suggested by SHR as well as our system prediction in green if it falls in SHR's candidate space, otherwise in orange.}
    \label{figure:SHR}
\end{figure*}
\begin{figure*}[!tbh]
    \centering
      \subfigure[]{\includegraphics[width=0.45\textwidth]{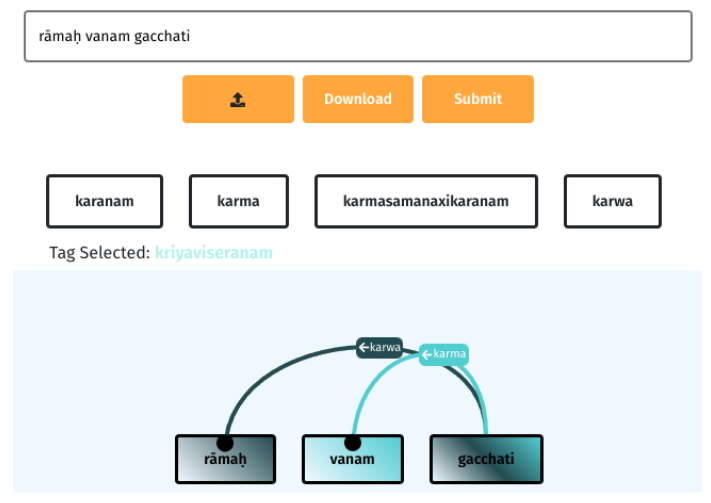}\label{fig:dep_parser}}
       \subfigure[]{\includegraphics[width=0.45\textwidth]{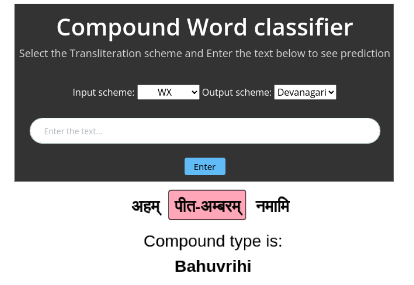}\label{fig:sacti}} 
  
\caption{ (a) Dependency parser: Interactive module for the dependency parsing task which directly loads predicted dependency trees from our pretrain model and allows user to correct mispredictions using our interactive interface. (b) Illustration of compound identifier}
    \label{figure:pipeline}
\end{figure*}
\section{Sanskrit Resources in SanskritShala}
\label{embed_dataset}
In this section, we describe  7 word embeddings pretrained on Sanskrit corpora and suit of 4 intrinsic tasks datasets to assess the quality of word embeddings, followed by  the description of  web interface.

\paragraph{Pretrained word embeddings for Sanskrit:}
There are two types of embedding methods: static and contextualized. Table \ref{tab:introTable} shows how they are categorized based on the smallest unit of input to the embedding model, such as character, subword, or token level. The paper focuses on two token-level word embeddings: \newcite[word2vec]{mikolov2013distributed} and \newcite[GloVe]{pennington-etal-2014-glove}. Word2vec is the foundation for all subsequent embeddings and works on the local context window, while GloVe considers the global context.
\begin{table}[!h]
\centering
\resizebox{0.45\textwidth}{!}{%
\begin{tabular}{|c|c|c|}
\hline
\textbf{Class}  & \textbf{Input type} & \textbf{Systems}    \\ \hline
Static   & character & charLM    \\ \hline 
   & subword & fastText    \\ \hline
   & token & word2vec, gloVe, LCM    \\ \hline\hline

Contextualized   & character & ELMo    \\ \hline
                & subword & ALBERT   \\ \hline 

\end{tabular}%
}
\caption{Overview of Sanskrit pretrained embeddings.}
\label{tab:introTable}
\end{table}
To address the OOV issue, subword \cite{wieting-etal-2016-charagram,bojanowski-etal-2017-enriching,heinzerling-strube-2018-bpemb} and character-level \cite{kim_charLM,jozefowicz2016exploring}  modeling have been proposed. 
We also explore two contextualized embeddings: ELMo \cite{elmo} and ALBERT \cite{ALBERT:}, a lighter version of BERT. 
We trained these 6 embedding methods on Sanskrit corpora and made the pretrained models publicly available \cite{sandhan-etal-2023-evaluating}.\footnote{\url{https://github.com/Jivnesh/SanskritShala/tree/master/EvalSan}}  The following section describes our proposed pretraining for low-resource settings.
\paragraph{LCM Pretraining:}
 We propose a supervised pretraining, which automatically leverages morphological information using the pretrained encoders.
In a nutshell, LCM integrates word representations from multiple encoders trained on three independent auxiliary tasks into the encoder of the neural dependency parser. LCM is acronym for three auxiliary tasks: Lemma prediction, Case prediction and Morphological tag prediction. LCM follows a pipeline-based approach consisting of two steps: pretraining and integration. Pretraining uses a sequence labelling paradigm and trains encoders for three independent auxiliary tasks. Later, these pretrained encoders are combined with the encoder of the neural parser via a gating mechanism similar to ~\newcite{sato-etal-2017-adversarial}. The LCM consists of three sequence labelling-based auxiliary tasks, namely, predicting the dependency label between a modifier-modified pair (\textbf{LT}), the monolithic morphological label \textbf{(MT)}, and the case attribute of each nominal \textbf{(CT)}.
 We encourage readers to refer \newcite[LCM]{sandhan-etal-2021-little} for more details.

 \paragraph{Datasets:}
The quality of word embedding spaces is evaluated through intrinsic and extrinsic methods. This study focuses on intrinsic evaluation, which involves assessing semantic and syntactic information in the words without testing on NLP applications. It is based on works such as \newcite{mikolov2013distributed} and \newcite{baroni-etal-2014-dont}.
These evaluations require a query inventory containing a query word and a related target word. However, such query inventories are not readily available for Sanskrit. To address this, we annotated query inventories for 4 intrinsic tasks: analogy prediction, synonym detection, relatedness, and concept categorization. The inventories were constructed using resources such as Sanskrit WordNet \cite{Kulkarni2017}, Amarako\d{s}a  \cite{inproceedings_nair}, and Sanskrit Heritage Reader \cite{goyaldesign16,heut_13}. 
 
\paragraph{Web Interface:}
\label{webinterface}
Figure \ref{figure:sanskritshala} shows our SanskritShala toolkit that offers interactive web-based predictions for various NLP tasks. The toolkit is built using React framework, which makes it user-friendly and easy to use. One of the tasks it handles is the word segmentation task, which is built on top of the web-based application called SHR++. The SHR++ demonstration is depicted in Figure \ref{fig:shr++}. The user inputs a Sanskrit string, which is then sent in real-time to SHR for potential word splits. The system prediction is then obtained from the pretrained word tokenizer. The human annotator is presented with the candidate solution space, with the system prediction highlighted in yellow.
The toolkit also features a flask-based application for morphological tagging, which takes user input and scrapes possible morphological tags for each word using SHR. As shown in Figure \ref{fig:morph_tagger}, the predictions of the pretrained morphological tagger are displayed in green or orange, depending on whether they are present in the candidate solution of SHR or not. The user can also add a new tag if the actual tag is missing in the SHR solution space or the system's prediction.
For the dependency parsing module, we have built a react-based front-end. The user input is passed to the pretrained model to generate a dependency structure. As illustrated in Figure \ref{fig:dep_parser}, the front-end automatically loads the predicted dependency tree and allows the user to make corrections if there are any mispredictions. Additionally, Figure \ref{fig:sacti} shows a flask-based application for the compound type identifier, where users can give input to the system through its web interface. The final annotations can be downloaded after each individual module. We plan to maintain the progress of Sanskrit NLP and offer an overview of available datasets and existing state-of-the-art via the leaderboard for various tasks.

\paragraph{Interactive Chatbot:}
SanskritShala-bot is a rule-based chatbot that makes it easy to automate simple and repetitive user requests, like answering frequently asked questions and directing users to relevant resources. It is also easier to set up and maintain than AI-powered chatbots, which are more complicated. SanskritShala-bot is useful for addressing frequently asked standard queries. It helps familiarize users with the platform by providing them with information and guidance on how to use it. It can answer questions about the platform's features, help users find their way around it, and explain step-by-step how to do certain tasks. This can make it easier for users to get started and  leading to a better user experience. 

\section{Conclusion}
We present the first neural-based Sanskrit NLP toolkit, SanskritShala which facilitates diverse linguistic analysis for tasks such as word segmentation, morphological tagging, dependency parsing and compound type identification. It is set up as a web-based application to make the toolkit easier to use for teaching and annotating. All the codebase, datasets and web-based applications are publicly available. We also release word embedding models trained on publicly available Sanskrit corpora and various annotated datasets for 4 intrinsic evaluation tasks to assess the intrinsic properties of word embeddings. We strongly believe that our toolkit will benefit people who are willing to work with Sanskrit and will eventually accelerate the Sanskrit NLP research.

\section*{Limitations}
We plan to extend SanskritShala  by integrating more downstream tasks such as Post-OCR correction, named entity recognition, verse recommendation, word order linearisation, and machine translation.
Improving the performance of existing tasks would be important. For example, the current dependency parser is very fragile (performance drops by 50\%) in the poetry domain.

\section*{Ethics Statement}
Our work involves the development of a platform for annotating Sanskrit text. We believe that this platform will be useful for people who are willing to work with Sanskrit for research and educational purposes. We have ensured that our platform is designed ethically and responsibly.
We do not foresee any harmful effects of our platform on any community. However, we caution users to use the platform carefully as our pretrained models are not perfect, and errors can occur in the annotation process.
All our systems are built using publicly available benchmark datasets, and we have released all our pretrained models and source codes publicly for future research. We are committed to transparency and open access in our work, and we believe that sharing our resources will benefit the wider NLP community.
We also acknowledge that NLP research can have potential ethical implications, particularly in areas such as data privacy, bias and discrimination. We are committed to continuing to consider these ethical implications as we develop our platform, and we welcome feedback from the community on how we can improve our ethical practices.

\section*{Acknowledgements}
We are thankful to Oliver Hellwig for the
DCS dataset and Gerard Huet for the Sanskrit Heritage Engine. We are grateful to Hackathon organizers\footnote{\url{https://sanskritpanini.github.io/index.html}} who encouraged us to build the best performing morphological tagger. We thank Amrith Krishna, Uniphore for providing the SHR++ interface as a starting point for the web interface of SanskritShala. We appreciate the assistance of Bishal Santra and Suman Chakraborty, IIT Kharagpur in deploying SanskritShala on the IIT Kharagpur server. We are grateful to Hritik Sharma, IIT Kanpur for helping us build a React-based front-end for SanskritShala.
We are thankful to Hrishikesh Terdalkar, IIT Kanpur for helpful discussions on deploying systems.
We appreciate the anonymous reviewers' insightful suggestions for enhancing this work. We'd like to say thanks to everyone who helped us make the different neural models for SanskritShala. The work was supported in part by the National Language Translation Mission (NLTM): Bhashini project by Government of India. The work of the first author is supported by the TCS Fellowship under the Project
TCS/EE/2011191P.

\bibliography{anthology,custom}

\begin{thebibliography}{42}
\expandafter\ifx\csname natexlab\endcsname\relax\def\natexlab#1{#1}\fi

\bibitem[{Adiga et~al.(2021)Adiga, Kumar, Krishna, Jyothi, Ramakrishnan, and
  Goyal}]{adiga-etal-2021-automatic}
Devaraja Adiga, Rishabh Kumar, Amrith Krishna, Preethi Jyothi, Ganesh
  Ramakrishnan, and Pawan Goyal. 2021.
\newblock \href {https://doi.org/10.18653/v1/2021.findings-acl.447} {Automatic
  speech recognition in {S}anskrit: A new speech corpus and modelling
  insights}.
\newblock In \emph{Findings of the Association for Computational Linguistics:
  ACL-IJCNLP 2021}, pages 5039--5050, Online. Association for Computational
  Linguistics.

\bibitem[{Baroni et~al.(2014)Baroni, Dinu, and
  Kruszewski}]{baroni-etal-2014-dont}
Marco Baroni, Georgiana Dinu, and Germ{\'a}n Kruszewski. 2014.
\newblock \href {https://doi.org/10.3115/v1/P14-1023} {Don{'}t count, predict!
  a systematic comparison of context-counting vs. context-predicting semantic
  vectors}.
\newblock In \emph{Proceedings of the 52nd Annual Meeting of the Association
  for Computational Linguistics (Volume 1: Long Papers)}, pages 238--247,
  Baltimore, Maryland. Association for Computational Linguistics.

\bibitem[{Bojanowski et~al.(2017)Bojanowski, Grave, Joulin, and
  Mikolov}]{bojanowski-etal-2017-enriching}
Piotr Bojanowski, Edouard Grave, Armand Joulin, and Tomas Mikolov. 2017.
\newblock \href {https://doi.org/10.1162/tacl_a_00051} {Enriching word vectors
  with subword information}.
\newblock \emph{Transactions of the Association for Computational Linguistics},
  5:135--146.

\bibitem[{Goyal et~al.(2012)Goyal, Huet, Kulkarni, Scharf, and
  Bunker}]{goyal-etal-2012-distributed}
Pawan Goyal, G{\'e}rard Huet, Amba Kulkarni, Peter Scharf, and Ralph Bunker.
  2012.
\newblock \href {https://aclanthology.org/C12-1062} {A distributed platform for
  {S}anskrit processing}.
\newblock In \emph{Proceedings of {COLING} 2012}, pages 1011--1028, Mumbai,
  India. The COLING 2012 Organizing Committee.

\bibitem[{Goyal and Huet(2016{\natexlab{a}})}]{goyal2016}
Pawan Goyal and Gérard Huet. 2016{\natexlab{a}}.
\newblock \href {https://doi.org/10.15398/jlm.v4i2.108} {Design and analysis of
  a lean interface for sanskrit corpus annotation}.
\newblock \emph{Journal of Language Modelling}, 4:145.

\bibitem[{Goyal and Huet(2016{\natexlab{b}})}]{goyaldesign16}
Pawan Goyal and Gérard Huet. 2016{\natexlab{b}}.
\newblock \href {https://doi.org/10.15398/jlm.v4i2.108} {Design and analysis of
  a lean interface for sanskrit corpus annotation}.
\newblock \emph{Journal of Language Modelling}, 4:145.

\bibitem[{Gupta et~al.(2020)Gupta, Krishna, Goyal, and
  Hellwig}]{gupta-etal-2020-evaluating}
Ashim Gupta, Amrith Krishna, Pawan Goyal, and Oliver Hellwig. 2020.
\newblock \href {https://doi.org/10.18653/v1/2020.sigmorphon-1.23} {Evaluating
  neural morphological taggers for {S}anskrit}.
\newblock In \emph{Proceedings of the 17th SIGMORPHON Workshop on Computational
  Research in Phonetics, Phonology, and Morphology}, pages 198--203, Online.
  Association for Computational Linguistics.

\bibitem[{Heinzerling and Strube(2018)}]{heinzerling-strube-2018-bpemb}
Benjamin Heinzerling and Michael Strube. 2018.
\newblock \href {https://aclanthology.org/L18-1473} {{BPE}mb: Tokenization-free
  pre-trained subword embeddings in 275 languages}.
\newblock In \emph{Proceedings of the Eleventh International Conference on
  Language Resources and Evaluation ({LREC} 2018)}, Miyazaki, Japan. European
  Language Resources Association (ELRA).

\bibitem[{Hellwig and Nehrdich(2018)}]{hellwig-nehrdich-2018-sanskrit}
Oliver Hellwig and Sebastian Nehrdich. 2018.
\newblock \href {https://doi.org/10.18653/v1/D18-1295} {{S}anskrit word
  segmentation using character-level recurrent and convolutional neural
  networks}.
\newblock In \emph{Proceedings of the 2018 Conference on Empirical Methods in
  Natural Language Processing}, pages 2754--2763, Brussels, Belgium.
  Association for Computational Linguistics.

\bibitem[{Huet and Goyal(2013)}]{heut_13}
Gérard Huet and Pawan Goyal. 2013.
\newblock Design of a lean interface for sanskrit corpus annotation.

\bibitem[{Jozefowicz et~al.(2016)Jozefowicz, Vinyals, Schuster, Shazeer, and
  Wu}]{jozefowicz2016exploring}
Rafal Jozefowicz, Oriol Vinyals, Mike Schuster, Noam Shazeer, and Yonghui Wu.
  2016.
\newblock \href {http://arxiv.org/abs/1602.02410} {Exploring the limits of
  language modeling}.

\bibitem[{Kim et~al.(2016)Kim, Jernite, Sontag, and Rush}]{kim_charLM}
Yoon Kim, Yacine Jernite, David Sontag, and Alexander~M. Rush. 2016.
\newblock Character-aware neural language models.
\newblock In \emph{Proceedings of the Thirtieth AAAI Conference on Artificial
  Intelligence}, AAAI'16, page 2741–2749. AAAI Press.

\bibitem[{Kondratyuk et~al.(2018)Kondratyuk, Gaven{\v{c}}iak, Straka, and
  Haji{\v{c}}}]{kondratyuk-etal-2018-lemmatag}
Daniel Kondratyuk, Tom{\'a}{\v{s}} Gaven{\v{c}}iak, Milan Straka, and Jan
  Haji{\v{c}}. 2018.
\newblock \href {https://doi.org/10.18653/v1/D18-1532} {{L}emma{T}ag: Jointly
  tagging and lemmatizing for morphologically rich languages with {BRNN}s}.
\newblock In \emph{Proceedings of the 2018 Conference on Empirical Methods in
  Natural Language Processing}, pages 4921--4928, Brussels, Belgium.
  Association for Computational Linguistics.

\bibitem[{Krishna et~al.(2023)Krishna, Gupta, Garasangi, Sandhan, Satuluri, and
  Goyal}]{krishna-etal-2023-neural}
Amrith Krishna, Ashim Gupta, Deepak Garasangi, Jeevnesh Sandhan, Pavankumar
  Satuluri, and Pawan Goyal. 2023.
\newblock \href {https://aclanthology.org/2023.wsc-csdh.1} {Neural approaches
  for data driven dependency parsing in {S}anskrit}.
\newblock In \emph{Proceedings of the Computational {S}anskrit {\&} Digital
  Humanities: Selected papers presented at the 18th World {S}anskrit
  Conference}, pages 1--20, Canberra, Australia (Online mode). Association for
  Computational Linguistics.

\bibitem[{Krishna et~al.(2020{\natexlab{a}})Krishna, Gupta, Garasangi,
  Satuluri, and Goyal}]{krishna-etal-2020-keep}
Amrith Krishna, Ashim Gupta, Deepak Garasangi, Pavankumar Satuluri, and Pawan
  Goyal. 2020{\natexlab{a}}.
\newblock \href {https://doi.org/10.18653/v1/2020.emnlp-main.388} {Keep it
  surprisingly simple: A simple first order graph based parsing model for joint
  morphosyntactic parsing in {S}anskrit}.
\newblock In \emph{Proceedings of the 2020 Conference on Empirical Methods in
  Natural Language Processing (EMNLP)}, pages 4791--4797, Online. Association
  for Computational Linguistics.

\bibitem[{Krishna et~al.(2018)Krishna, Santra, Bandaru, Sahu, Sharma, Satuluri,
  and Goyal}]{krishna-etal-2018-free}
Amrith Krishna, Bishal Santra, Sasi~Prasanth Bandaru, Gaurav Sahu, Vishnu~Dutt
  Sharma, Pavankumar Satuluri, and Pawan Goyal. 2018.
\newblock \href {https://doi.org/10.18653/v1/D18-1276} {Free as in free word
  order: An energy based model for word segmentation and morphological tagging
  in {S}anskrit}.
\newblock In \emph{Proceedings of the 2018 Conference on Empirical Methods in
  Natural Language Processing}, pages 2550--2561, Brussels, Belgium.
  Association for Computational Linguistics.

\bibitem[{Krishna et~al.(2020{\natexlab{b}})Krishna, Santra, Gupta, Satuluri,
  and Goyal}]{krishna-etal-2020-graph}
Amrith Krishna, Bishal Santra, Ashim Gupta, Pavankumar Satuluri, and Pawan
  Goyal. 2020{\natexlab{b}}.
\newblock \href {https://doi.org/10.1162/coli_a_00390} {A graph-based framework
  for structured prediction tasks in {S}anskrit}.
\newblock \emph{Computational Linguistics}, 46(4):785--845.

\bibitem[{Krishna et~al.(2021)Krishna, Santra, Gupta, Satuluri, and
  Goyal}]{Graph-Based}
Amrith Krishna, Bishal Santra, Ashim Gupta, Pavankumar Satuluri, and Pawan
  Goyal. 2021.
\newblock \href {https://doi.org/10.1162/coli_a_00390} {{A Graph-Based
  Framework for Structured Prediction Tasks in Sanskrit}}.
\newblock \emph{Computational Linguistics}, 46(4):785--845.

\bibitem[{Krishna et~al.(2016)Krishna, Satuluri, Sharma, Kumar, and
  Goyal}]{krishna-etal-2016-compound}
Amrith Krishna, Pavankumar Satuluri, Shubham Sharma, Apurv Kumar, and Pawan
  Goyal. 2016.
\newblock \href {https://aclanthology.org/W16-3701} {Compound type
  identification in {S}anskrit: What roles do the corpus and grammar play?}
\newblock In \emph{Proceedings of the 6th Workshop on South and Southeast
  {A}sian Natural Language Processing ({WSSANLP}2016)}, pages 1--10, Osaka,
  Japan. The COLING 2016 Organizing Committee.

\bibitem[{Krishna et~al.(2019)Krishna, Sharma, Santra, Chakraborty, Satuluri,
  and Goyal}]{krishna-etal-2019-poetry}
Amrith Krishna, Vishnu Sharma, Bishal Santra, Aishik Chakraborty, Pavankumar
  Satuluri, and Pawan Goyal. 2019.
\newblock \href {https://doi.org/10.18653/v1/P19-1111} {Poetry to prose
  conversion in {S}anskrit as a linearisation task: A case for low-resource
  languages}.
\newblock In \emph{Proceedings of the 57th Annual Meeting of the Association
  for Computational Linguistics}, pages 1160--1166, Florence, Italy.
  Association for Computational Linguistics.

\bibitem[{Krishna et~al.(2020{\natexlab{c}})Krishna, Vidhyut, Chawla, Sambhavi,
  and Goyal}]{krishna-etal-2020-shr}
Amrith Krishna, Shiv Vidhyut, Dilpreet Chawla, Sruti Sambhavi, and Pawan Goyal.
  2020{\natexlab{c}}.
\newblock \href {https://aclanthology.org/2020.lrec-1.874} {{SHR}++: An
  interface for morpho-syntactic annotation of {S}anskrit corpora}.
\newblock In \emph{Proceedings of the Twelfth Language Resources and Evaluation
  Conference}, pages 7069--7076, Marseille, France. European Language Resources
  Association.

\bibitem[{Krishnan et~al.(2020)Krishnan, Kulkarni, and Huet}]{hackathon}
Sriram Krishnan, Amba Kulkarni, and Gérard Huet. 2020.
\newblock \href {https://doi.org/10.48550/ARXIV.2005.06545} {Validation and
  normalization of dcs corpus using sanskrit heritage tools to build a tagged
  gold corpus}.

\bibitem[{Kulkarni et~al.(2020)Kulkarni, Satuluri, Panchal, Maity, and
  Malvade}]{kulkarni-etal-2020-dependency}
Amba Kulkarni, Pavankumar Satuluri, Sanjeev Panchal, Malay Maity, and Amruta
  Malvade. 2020.
\newblock \href {https://doi.org/10.18653/v1/2020.tlt-1.12} {Dependency
  relations for {S}anskrit parsing and treebank}.
\newblock In \emph{Proceedings of the 19th International Workshop on Treebanks
  and Linguistic Theories}, pages 135--150, D{\"u}sseldorf, Germany.
  Association for Computational Linguistics.

\bibitem[{Kulkarni and Sharma(2019)}]{kulkarni-sharma-2019-paninian}
Amba Kulkarni and Dipti Sharma. 2019.
\newblock \href {https://doi.org/10.18653/v1/W19-7724} {{P}{\=a}ṇinian
  syntactico-semantic relation labels}.
\newblock In \emph{Proceedings of the Fifth International Conference on
  Dependency Linguistics (Depling, SyntaxFest 2019)}, pages 198--208, Paris,
  France. Association for Computational Linguistics.

\bibitem[{Kulkarni(2017)}]{Kulkarni2017}
Malhar Kulkarni. 2017.
\newblock \href {https://doi.org/10.1007/978-981-10-1909-8_14} {\emph{Sanskrit
  WordNet at Indian Institute of Technology (IITB) Mumbai}}, pages 231--241.
  Springer Singapore, Singapore.

\bibitem[{Lan et~al.(2020)Lan, Chen, Goodman, Gimpel, Sharma, and
  Soricut}]{ALBERT:}
Zhenzhong Lan, Mingda Chen, Sebastian Goodman, Kevin Gimpel, Piyush Sharma, and
  Radu Soricut. 2020.
\newblock \href {https://openreview.net/forum?id=H1eA7AEtvS} {Albert: A lite
  bert for self-supervised learning of language representations}.
\newblock In \emph{International Conference on Learning Representations}.

\bibitem[{Mikolov et~al.(2013)Mikolov, Sutskever, Chen, Corrado, and
  Dean}]{mikolov2013distributed}
Tomas Mikolov, Ilya Sutskever, Kai Chen, Greg~S Corrado, and Jeff Dean. 2013.
\newblock Distributed representations of words and phrases and their
  compositionality.
\newblock \emph{Advances in neural information processing systems},
  26:3111--3119.

\bibitem[{Nair and Kulkarni(2010)}]{inproceedings_nair}
Sivaja Nair and Amba Kulkarni. 2010.
\newblock \href {https://doi.org/10.1007/978-3-642-17528-2_13} {The knowledge
  structure in amarakosa.}
\newblock pages 173--189.

\bibitem[{Pennington et~al.(2014)Pennington, Socher, and
  Manning}]{pennington-etal-2014-glove}
Jeffrey Pennington, Richard Socher, and Christopher Manning. 2014.
\newblock \href {https://doi.org/10.3115/v1/D14-1162} {{G}lo{V}e: Global
  vectors for word representation}.
\newblock In \emph{Proceedings of the 2014 Conference on Empirical Methods in
  Natural Language Processing ({EMNLP})}, pages 1532--1543, Doha, Qatar.
  Association for Computational Linguistics.

\bibitem[{Peters et~al.(2018)Peters, Neumann, Iyyer, Gardner, Clark, Lee, and
  Zettlemoyer}]{elmo}
Matthew Peters, Mark Neumann, Mohit Iyyer, Matt Gardner, Christopher Clark,
  Kenton Lee, and Luke Zettlemoyer. 2018.
\newblock \href {https://doi.org/10.18653/v1/N18-1202} {Deep contextualized
  word representations}.
\newblock In \emph{Proceedings of the 2018 Conference of the North {A}merican
  Chapter of the Association for Computational Linguistics: Human Language
  Technologies, Volume 1 (Long Papers)}, pages 2227--2237, New Orleans,
  Louisiana. Association for Computational Linguistics.

\bibitem[{Sandhan et~al.(2022{\natexlab{a}})Sandhan, Behera, and
  Goyal}]{sandhan_systematic}
Jivnesh Sandhan, Laxmidhar Behera, and Pawan Goyal. 2022{\natexlab{a}}.
\newblock \href {https://doi.org/10.48550/ARXIV.2201.11374} {Systematic
  investigation of strategies tailored for low-resource settings for sanskrit
  dependency parsing}.

\bibitem[{Sandhan et~al.(2022{\natexlab{b}})Sandhan, Daksh, Paranjay, Behera,
  and Goyal}]{sandhan-etal-2022-prabhupadavani}
Jivnesh Sandhan, Ayush Daksh, Om~Adideva Paranjay, Laxmidhar Behera, and Pawan
  Goyal. 2022{\natexlab{b}}.
\newblock \href {https://aclanthology.org/2022.latechclfl-1.4} {Prabhupadavani:
  A code-mixed speech translation data for 25 languages}.
\newblock In \emph{Proceedings of the 6th Joint SIGHUM Workshop on
  Computational Linguistics for Cultural Heritage, Social Sciences, Humanities
  and Literature}, pages 24--29, Gyeongju, Republic of Korea. International
  Conference on Computational Linguistics.

\bibitem[{Sandhan et~al.(2022{\natexlab{c}})Sandhan, Gupta, Terdalkar, Sandhan,
  Samanta, Behera, and Goyal}]{sandhan-etal-2022-novel}
Jivnesh Sandhan, Ashish Gupta, Hrishikesh Terdalkar, Tushar Sandhan, Suvendu
  Samanta, Laxmidhar Behera, and Pawan Goyal. 2022{\natexlab{c}}.
\newblock \href {https://aclanthology.org/2022.coling-1.358} {A novel
  multi-task learning approach for context-sensitive compound type
  identification in {S}anskrit}.
\newblock In \emph{Proceedings of the 29th International Conference on
  Computational Linguistics}, pages 4071--4083, Gyeongju, Republic of Korea.
  International Committee on Computational Linguistics.

\bibitem[{Sandhan et~al.(2019)Sandhan, Krishna, Goyal, and
  Behera}]{sandhan-etal-2019-revisiting}
Jivnesh Sandhan, Amrith Krishna, Pawan Goyal, and Laxmidhar Behera. 2019.
\newblock \href {https://aclanthology.org/W19-7503} {Revisiting the role of
  feature engineering for compound type identification in {S}anskrit}.
\newblock In \emph{Proceedings of the 6th International Sanskrit Computational
  Linguistics Symposium}, pages 28--44, IIT Kharagpur, India. Association for
  Computational Linguistics.

\bibitem[{Sandhan et~al.(2021)Sandhan, Krishna, Gupta, Behera, and
  Goyal}]{sandhan-etal-2021-little}
Jivnesh Sandhan, Amrith Krishna, Ashim Gupta, Laxmidhar Behera, and Pawan
  Goyal. 2021.
\newblock \href {https://doi.org/10.18653/v1/2021.eacl-srw.16} {A little
  pretraining goes a long way: A case study on dependency parsing task for
  low-resource morphologically rich languages}.
\newblock In \emph{Proceedings of the 16th Conference of the European Chapter
  of the Association for Computational Linguistics: Student Research Workshop},
  pages 111--120, Online. Association for Computational Linguistics.

\bibitem[{Sandhan et~al.(2023)Sandhan, Paranjay, Digumarthi, Behra, and
  Goyal}]{sandhan-etal-2023-evaluating}
Jivnesh Sandhan, Om~Adideva Paranjay, Komal Digumarthi, Laxmidhar Behra, and
  Pawan Goyal. 2023.
\newblock \href {https://aclanthology.org/2023.wsc-csdh.2} {Evaluating neural
  word embeddings for {S}anskrit}.
\newblock In \emph{Proceedings of the Computational {S}anskrit {\&} Digital
  Humanities: Selected papers presented at the 18th World {S}anskrit
  Conference}, pages 21--37, Canberra, Australia (Online mode). Association for
  Computational Linguistics.

\bibitem[{Sandhan et~al.(2022{\natexlab{d}})Sandhan, Singha, Rao, Samanta,
  Behera, and Goyal}]{translist}
Jivnesh Sandhan, Rathin Singha, Narein Rao, Suvendu Samanta, Laxmidhar Behera,
  and Pawan Goyal. 2022{\natexlab{d}}.
\newblock \href {https://doi.org/10.48550/ARXIV.2210.11753} {Translist: A
  transformer-based linguistically informed sanskrit tokenizer}.

\bibitem[{Sato et~al.(2017)Sato, Manabe, Noji, and
  Matsumoto}]{sato-etal-2017-adversarial}
Motoki Sato, Hitoshi Manabe, Hiroshi Noji, and Yuji Matsumoto. 2017.
\newblock \href {https://doi.org/10.18653/v1/K17-3007} {Adversarial training
  for cross-domain {U}niversal {D}ependency parsing}.
\newblock In \emph{Proceedings of the {C}o{NLL} 2017 Shared Task: Multilingual
  Parsing from Raw Text to Universal Dependencies}, pages 71--79, Vancouver,
  Canada. Association for Computational Linguistics.

\bibitem[{Sriram et~al.(2023)Sriram, Kulkarni, and
  Huet}]{sriram-etal-2023-validation}
Krishnan Sriram, Amba Kulkarni, and G{\'e}rard Huet. 2023.
\newblock \href {https://aclanthology.org/2023.wsc-csdh.3} {Validation and
  normalization of {DCS} corpus and development of the {S}anskrit heritage
  engine{'}s segmenter}.
\newblock In \emph{Proceedings of the Computational {S}anskrit {\&} Digital
  Humanities: Selected papers presented at the 18th World {S}anskrit
  Conference}, pages 38--58, Canberra, Australia (Online mode). Association for
  Computational Linguistics.

\bibitem[{Terdalkar and Bhattacharya(2021)}]{Sangrahaka}
Hrishikesh Terdalkar and Arnab Bhattacharya. 2021.
\newblock \href {https://doi.org/10.1145/3468264.3473113} {Sangrahaka: A tool
  for annotating and querying knowledge graphs}.
\newblock In \emph{Proceedings of the 29th ACM Joint Meeting on European
  Software Engineering Conference and Symposium on the Foundations of Software
  Engineering}, ESEC/FSE 2021, page 1520–1524, New York, NY, USA. Association
  for Computing Machinery.

\bibitem[{Terdalkar and Bhattacharya(2022)}]{Chandojnanam}
Hrishikesh Terdalkar and Arnab Bhattacharya. 2022.
\newblock \href {https://doi.org/10.48550/ARXIV.2209.14924} {Chandojnanam: A
  sanskrit meter identification and utilization system}.

\bibitem[{Wieting et~al.(2016)Wieting, Bansal, Gimpel, and
  Livescu}]{wieting-etal-2016-charagram}
John Wieting, Mohit Bansal, Kevin Gimpel, and Karen Livescu. 2016.
\newblock \href {https://doi.org/10.18653/v1/D16-1157} {{C}haragram: Embedding
  words and sentences via character n-grams}.
\newblock In \emph{Proceedings of the 2016 Conference on Empirical Methods in
  Natural Language Processing}, pages 1504--1515, Austin, Texas. Association
  for Computational Linguistics.

\end{thebibliography}
\bibliographystyle{acl_natbib}

\appendix



\end{document}